\documentclass[conference]{IEEEtran}

\usepackage{bbm}
\usepackage{adjustbox}
\usepackage{array}
\usepackage[table]{xcolor}
\usepackage{booktabs}%
\usepackage{caption} 
\usepackage{soul}
\usepackage{url}
\usepackage[linesnumbered,ruled,vlined]{algorithm2e}
\usepackage{setspace}
\usepackage{hyperref}
\SetKwInput{KwInput}{Input}
\SetKwInput{KwOutput}{Output}

\usepackage{amsmath}
\usepackage{amsfonts}
\usepackage{multirow}
\DeclareMathOperator*{\argmax}{arg\,max}
\DeclareMathOperator*{\argmin}{arg\,min}

\usepackage{pifont}
\usepackage{subfig}

\definecolor{red_pastel}{RGB}{210, 42, 42} 
\definecolor{blue_pastel}{RGB}{4, 54, 183} 
\definecolor{blue_mpl}{RGB}{31, 119, 180} 
\definecolor{orange_mpl}{RGB}{255, 165, 0} 

\begin{document}

\title{
Early Classification of Time Series in \\ Non-Stationary Cost Regimes 
}

\author{\IEEEauthorblockN{Aurélien Renault}
 \IEEEauthorblockA{Orange Research, France\\
 Email: aurelien.renault@orange.com} \\
 \IEEEauthorblockN{Antoine Cornuéjols}
 \IEEEauthorblockA{AgroParisTech, France\\
 Email: antoine.cornuejols@agroparistech.fr} \\
 \and
 \IEEEauthorblockN{Alexis Bondu}
 \IEEEauthorblockA{Orange Research, France\\
 Email: alexis.bondu@orange.com}\\ 
 \IEEEauthorblockN{Vincent Lemaire}
 \IEEEauthorblockA{Orange Research, France\\ 
 Email: vincent.lemaire@orange.com}}

\maketitle

\begin{abstract}

Early Classification of Time Series (ECTS) addresses decision-making problems in which predictions must be made as early as possible while maintaining high accuracy. Most existing ECTS methods assume that the time-dependent decision costs governing the learning objective are known, fixed, and correctly specified. In practice, however, these costs are often uncertain and may change over time, leading to mismatches between training-time and deployment-time objectives. In this paper, we study ECTS under two practically relevant forms of cost non-stationarity: drift in the balance between misclassification and decision delay costs, and stochastic realizations of decision costs that deviate from the nominal training-time model. To address these challenges, we revisit representative ECTS approaches and adapt them to an online learning setting. Focusing on separable methods, we update only the triggering model during deployment, while keeping the classifier fixed. We propose several online adaptations and baselines, including bandit-based and RL–based approaches, and conduct controlled experiments on synthetic data to systematically evaluate robustness under cost non-stationarity. Our results demonstrate that online learning can effectively improve the robustness of ECTS methods to cost drift, with RL–based strategies exhibiting strong and stable performance across varying cost regimes.

\end{abstract}

\section{Motivation}
\label{sec_motivation}

{Many real-world decision-making systems operate under \textit{time pressure}, where delaying a prediction can be as costly as making an incorrect one. Representative examples include real-time monitoring, predictive maintenance, medical triage, and adaptive human--machine interaction.
In such settings, the objective is not only to achieve high predictive accuracy, but to do so as early as possible. This inherent trade-off between classification accuracy and decision earliness lies at the core of \textit{Early Classification of Time Series} (ECTS) and has been extensively studied in this field \cite{gupta2020approaches, akasiadis2024framework, renaultearly}. 
ECTS belongs to the broader class of early decision-making \cite{bondu2022open} and anytime prediction problems.
More specifically, ECTS can be naturally formulated as an empirical risk minimization problem with a time-dependent loss function that jointly penalizes misclassification errors and decision delays (see Section \ref{sec_background}).}

Most existing ECTS approaches, however, rely on the assumption that this time-dependent loss and the decision costs it encodes are known, fixed, and accurately specified during training. 
In practice, these costs must be elicited from domain experts and are often subject to substantial uncertainty.
At deployment time, the costs actually incurred may deviate from those assumed during training due to imperfect cost specification, evolving operational conditions, or intrinsic randomness across decision instances. Although cost mismatches are inevitable in practice, their impact on ECTS systems remains largely unexplored, making cost non-stationarity a fundamental challenge for ECTS and, more broadly, for online and continual neural learning systems.
Concrete applications illustrate how decision costs may be non-stationary over time: 

{\textbf{I} - In healthcare monitoring \cite{Fleuren2020Meta}, for instance, the \textit{balance} between misclassification and decision delay costs evolves as clinical protocols, hospital workload, or available treatments change. Early detection of patient deterioration enables timely intervention; yet, premature alarms may lead to unnecessary treatments, inefficient resource allocation. Thus, the relative importance of accuracy versus earliness may drift over time, inducing a deployment-time objective that differs from the one assumed during training.}

{\textbf{II} - In addition to cost drift, \textit{stochastic} decision costs naturally arise in industrial early defect detection. Consider a production line manufacturing composite mechanical parts, such as carbon-fiber–reinforced components used in aerospace or automotive industries. Defective parts may be ejected from the line as scrap, but the economic cost of rejection increases as manufacturing progresses: early rejection allows partial material recovery, while late rejection often makes recycling infeasible once curing or assembly steps are completed. Crucially, the realized cost of a late decision depends on the type of defect—unknown at prediction time—which affects the ability to separate and reuse constituent materials. Thus, the actual decision cost is inherently stochastic and may deviate from the nominal cost assumed during training.}

{
%
Motivated by these observations, this paper provides a first systematic study of ECTS under two practically relevant forms of cost non-stationarity, which are explicitly addressed in our experimental evaluation:
(\textit{i}) drift in the balance between misclassification costs and decision delay costs, and (\textit{ii}) stochastic realizations of decision costs that deviate from the nominal training-time regime. To tackle these challenges, we revisit the ECTS literature with the goal of adapting widely used approaches to an online learning setting, a direction that has not been investigated to our knowledge. We focus on \textit{separable} ECTS methods, which first learn a classifier and subsequently learn a triggering model, as this paradigm encompasses the vast majority of existing work \cite{gupta2020approaches, akasiadis2024framework, renaultearly}. In our setting, only the triggering model is updated 
during deployment, while the classifier remains fixed, reflecting 
that cost drift primarily affects decision timing rather than class discrimination. Our contributions are threefold: }

\begin{itemize}

    \item {\textbf{Online adaptation of representative ECTS methods.} We adapt several widely used ECTS approaches to the online setting, including an online extension of \textit{Calimera} \cite{bilski2023calimera} based on a deep learning architecture. We also introduce (\textit{i}) four new baselines: two based on multi-armed bandits and two adapting the probability-thresholding paradigm commonly used in classical ECTS and (\textit{ii}) propose lightweight online adaptations of \textit{Alert} \cite{renault2025deep} and \textit{Economy} \cite{achenchabe2021early}, which were originally designed with online decision-making in mind but had not previously been evaluated in this context.}

    \item {\textbf{Controlled experiments on cost non-stationarity.} We present the first experimental evaluation of ECTS under both drift in the balance between misclassification and delay costs and stochastic cost realizations with controlled distributions, using synthetic data that enables precise control over deployment conditions. This evaluation framework readily extends to real-world applications.
    %
    }

    \item {\textbf{Proof of concept for cost-robust online ECTS.} Our results demonstrate that ECTS methods can be effectively adapted to online learning to improve robustness under non-stationary costs. In particular, we observe that RL–based approaches exhibit strong and stable behavior across varying cost regimes, paving the way for future real-world applications.}
        
\end{itemize}

The remainder of this paper is organized as follows. Section \ref{sec_background} provides background on ECTS. Section \ref{sec:related_works} reviews closely related research areas. Section \ref{sec_ects_online} details the proposed online formulation of ECTS and the methods introduced in this paper, including a categorization of the proposed approaches according to different update timings, corresponding to increasing levels of information available about a new instance at update time. Section \ref{sec_experimental_protocol} describes the experimental protocol, including the synthetic datasets, the controlled cost drift scenarios, and the evaluation metrics. Finally, Section \ref{sec_results} presents and discusses the experimental results for the two forms of cost non-stationarity considered in this work.


\section{Background}
\label{sec_background}

\subsection{Problem statement of standard ECTS}

{In the ECTS problem, an input time series of size $T$ is observed progressively. At time $t \leq T$, only a prefix ${\mathbf x}_t \, = \, \langle {x_1}, \ldots, {x_t} \rangle$ is available, where ${x_i}_{(1 \leq i \leq t)}$ denotes the measurements (possibly multivariate). The full series $\mathbf{x}_T$ belongs to an unknown class $y \in {\cal Y}$. The task is to output a prediction $\hat{y} \in {\cal Y}$ for the class of the incoming time series at some time $\hat{t} \in [1,T]$ before the deadline $T$.} 

{ECTS explicitly addresses a trade-off between predictive accuracy and earliness. The correctness of the prediction is measured by a misclassification cost $\mathrm{C}_m(\hat{y}|y)$, and time pressure is modeled by a delay cost $\mathrm{C}_d(t)$, assumed positive and typically non-decreasing over time. Formally, we consider:  $\mathrm{C}_m : \mathcal{Y} \times \mathcal{Y} \rightarrow \mathbb{R}, \quad \mathrm{C}_d : \mathbb{R}^+ \rightarrow \mathbb{R}$.} 

{An ECTS system involves a predictor $\hat{y}(\mathbf{x}_t)$ that outputs a class from any prefix $\mathbf{x}_t$. The cost incurred when a prediction is triggered at time $t$ is given by a time-dependent loss $\mathcal{L}(\hat{y}(\mathbf{x}_t), y, t) = \mathrm{C}_m(\hat{y}(\mathbf{x}_t) | y) + \mathrm{C}_d(t). $}

{If the full time series $\mathbf{x}_T$ and its label $y$ were known in advance, the optimal decision time would satisfy:} 

\begin{equation}
t^\star = \argmin_{t \in [1, T]} \mathcal{L}(\hat{y}({\mathbf x}_t), y, t).
\label{eq_optimal_time}
\end{equation}

{Let $s^\star \in \mathcal{S}$ be an optimal ECTS function belonging to a class of functions $\mathcal{S}$, whose output at time $t$ when receiving $\mathbf{x}_t$ is:} 

\begin{equation}
s^{\star}({\mathbf x}_t)  =  
\left\{
    \begin{array}{ll}
        \emptyset & \mbox{if extra measures are queried;}\\
        y^{\star} = \hat{y}(x_{t^{\star}}) & \mbox{if prediction is triggered at }t^{\star};\\    
    \end{array}
\right.
\label{eq:ects_model}
\end{equation}

{In practice, ECTS is an online optimization problem: at each time $t$, the function $s(\mathbf{x}_t)$ must decide whether to predict or wait for more data, based only on the prefix $\mathbf{x}_t$. Equation (\ref{eq_optimal_time}) is therefore not directly operational, as it assumes access to the full series. Instead, $s$ induces a stopping time $\hat{t}$ on each incoming series, and the goal is to choose $\hat{t}$ such that the incurred loss $\mathcal{L}(\hat{y}(\mathbf{x}_{\hat{t}}), y, \hat{t})$ is as small as possible.}

{From a machine learning perspective, the goal is to find a function $s \in {\cal S}$ that best optimizes the time-dependent loss function $\mathcal{L}$, minimizing the true risk over the data-generating process $\mathbb{P}{(\mathcal{X} \times \mathcal{Y})}$ \cite{renaultearly}:} 

\begin{equation}
        \argmin_{s \in \mathcal{S}} \mathbb{E}_{(\mathbf{x}, y) \sim \mathbb{P}_{(\mathcal{X} \times \mathcal{Y})}}\left[ {\mathcal{L}(\hat{y}({\mathbf x}_{\hat{t}}), y, \hat{t})} \right]
\label{eq:cost1}
\end{equation} 

{In order to solve the ECTS problem, a training set is made available that is composed of $N$ labeled time series, denoted by $({\mathbf x}^i_T, y^i)_{i \in [0, N]} \in (\mathcal{X} \times \mathcal{Y})$, 
where each series ${\mathbf x}_T = \, \langle {x_1}, \ldots, {x_T} \rangle $ is complete and of the same size $T$, and associated with its label $y \in \mathcal{Y}$.}

\subsection{Literature on standard ECTS}


{The ECTS literature can be broadly divided into \emph{end-to-end} and \emph{separable} approaches. End-to-end methods jointly learn a classifier and a triggering mechanism. While flexible, they are classifier-dependent and often difficult to compare consistently across studies. In contrast, \emph{separable} approaches decouple classification and decision triggering. This modular design dominates the literature and has been extensively evaluated in previous works \cite{gupta2020approaches,akasiadis2024framework,renaultearly}, which motivates the focus on separable approaches in this paper.}

A widely used separable baseline is \emph{Proba-Threshold}, which triggers a prediction when the maximum posterior probability exceeds a learned threshold. Despite its simplicity, this myopic confidence-based approach has been shown to achieve competitive performance across many datasets and cost settings, making it a natural reference method \cite{renaultearly}.

Beyond myopic strategies, non-myopic separable approaches explicitly anticipate future observations. The \emph{Economy} family \cite{achenchabe2021early, dachraoui2015early} triggers a decision when the expected cost at the current time step is minimal among all anticipated future ones, relying on estimates of future costs. \emph{Calimera} \cite{bilski2023calimera} follows a similar anticipation-based principle and uses regression models learned by backward induction to predict future costs. \emph{Alert} \cite{renault2025deep}, whose triggering mechanism is based on reinforcement learning, is inherently suited to online adaptation.  These methods consistently outperform purely myopic approaches while preserving a separable design \cite{renaultearly}.



{To the best of our knowledge, none of these ECTS approaches have been evaluated under deployment-time cost non-stationarity. In this work, \textit{Proba-Threshold}, \textit{Economy} and \textit{Calimera}  are adapted to the online learning setting, as these approaches provide strong performance and are widely used in the standard ECTS regime \cite{renaultearly, bilski2023calimera, renault2025deep}.}

\subsection{Online ECTS with deployment-time cost mismatch}

{The standard ECTS formulation assumes that the misclassification and delay costs used for training, $\mathrm{C}_m$ and $\mathrm{C}_d$, accurately reflect the costs incurred at deployment. In practice, this assumption is often violated: the \textit{actual} costs observed in production may differ from the nominal ones and can evolve over time. Such cost variations may occur either within the duration of a single time series (i.e. along the time index $t \in [0, T]$) or across longer time scales, for instance due to seasonal or operational changes.}

{To clearly distinguish these two temporal dimensions, we explicitly separate the time index of the time series from the deployment timeline. We denote by $\mathrm{C}^{(u)}_m$ and $\mathrm{C}^{(u)}_d$ the \textit{deployment-time} misclassification and delay costs at deployment step $u \in \mathbb{N}^+$, where $u$ indexes successive processed incoming-time series. We define the corresponding loss as follows:}
$\mathcal{L}^{(u)}(\hat{y}(\mathbf{x}_t), y, t) = \mathrm{C}^{(u)}_m(\hat{y}(\mathbf{x}_t) \mid y) + \mathrm{C}^{(u)}_d(t).$
{In the general case, $\mathrm{C}^{(u)}_m$ and $\mathrm{C}^{(u)}_d$  are \textit{(i)} \textit{unknown} to the learner at deployment; \textit{(ii)} may {differ} from the training-time costs; \textit{(iii)} and can be \textit{non-stationary} (drifting over time) and/or \textit{stochastic} (varying across instances).}



{At deployment, an ECTS system repeatedly processes incoming time series. For each new series, it chooses a prediction time $\hat{t}$ based on the observed prefix $\mathbf{x}_{\hat{t}}$, outputs a prediction $\hat{y}(\mathbf{x}_{\hat{t}})$, and then incurs a deployment-time loss $\mathcal{L}^{(u)}(\hat{y}(\mathbf{x}_{\hat{t}}), y, \hat{t})$ once the true label $y$ and the realized costs are revealed. Its deployment performance is measured by
\begin{equation}
\mathbb{E}\big[\mathcal{L}^{(u)}(\hat{y}(\mathbf{x}_{\hat{t}}), y, \hat{t})\big],
\label{eq:deployment-loss-short}
\end{equation}
where the expectation is taken over the data-generating process and, when applicable, over the randomness in the costs.}

{The purpose of this paper is to allow ECTS systems to \textit{adapt online} to these unknown deployment-time costs, while the classifier has been trained only once under nominal costs $\mathrm{C}_m, \mathrm{C}_d$. Concretely:
\begin{itemize}
    \item \textit{Offline}, a classifier is trained using complete labeled time series and nominal costs.
    \item \textit{Online}, during deployment, only the triggering model is updated based on the observed feedback $\mathcal{L}^{(u)}$, with the objective of minimizing the expected loss in Equation \ref{eq:deployment-loss-short} under cost drift and stochastic costs.
\end{itemize}
}

The remainder of the paper instantiates various online adaptation strategies for the most popular ECTS approaches and evaluates their robustness under controlled forms of cost non-stationarity.

\section{Related domains}
\label{sec:related_works}

This section reviews work related to the proposed setting of ECTS under deployment-time cost non-stationarity.

\subsection{Cost-Sensitive Learning}

{Learning algorithms that optimize decisions under user-defined cost functions have been extensively studied in cost-sensitive learning~\cite{elkan2001foundations,sheng2006thresholding,petrides2022cost}, typically by incorporating misclassification cost matrices into standard classifiers to minimize expected cost rather than the error rate (e.g., via threshold adjustment, cost-sensitive losses, or tailored ensembles).}

{In an online setting, cost-sensitive learning has also been explored, e.g., \cite{wang2013cost} integrate fixed misclassification costs into streaming classifiers; however, they do not address changes in the cost function over time.} 
{In contrast, the ECTS setting studied here optimizes an online time-dependent loss, combining misclassification and delay costs, under non-stationary deployment-time conditions.} 

\subsection{Concept Drift in Data Streams}

{Concept drift refers to changes in the joint distribution of inputs and labels over time, and is a central topic in data stream mining~\cite{harel2014concept,lu2018learning,iwashita2018overview}. Many works have studied how to detect, characterize, and adapt to drift in streaming environments. Existing approaches include drift detectors based on statistical tests or resampling~\cite{harel2014concept}, adaptive ensembles that reweight or replace base learners \cite{jaber2013new}, and prequential evaluation protocols that interleave prediction and update~\cite{lu2018learning}.}

{The scope of the present work is more specific. Rather than addressing changes in the data or label distribution, the focus is on \emph{cost drift}: the underlying data-generating process is assumed to be stationary, while the decision costs governing the loss function may evolve. From this perspective, the proposed setting can be seen as an instance of learning under a fixed concept but evolving evaluation criteria.}

\subsection{Non-Stationary Reinforcement Learning and Bandits}

{Online decision-making is a long-standing topic of study in artificial intelligence, which has led in particular to Reinforcement Learning (RL) algorithms and Multi-Armed Bandit (MAB) algorithms. In robust RL, the goal is to obtain policies that perform well under environment variations or distributional shifts, \cite{padakandla2020reinforcement,morimoto2005robust,moos2022robust}.}


{A closely related line of work studies non-stationary rewards for MAB, with algorithms such as discounted or sliding-window UCB~\cite{garivier2008upper} and associated regret bounds in drifting settings~\cite{besbes2014stochastic}, which trade off reactivity to change and robustness to noise via forgetting mechanisms.}

{The present work builds on these ideas by introducing bandit-based trigger models for ECTS, equipped with decay or window mechanisms to handle cost drift. These approaches are evaluated alongside supervised and RL–based methods, within controlled non-stationary cost regimes.}





\section{Bringing ECTS to Online Learning}
\label{sec_ects_online}

The trigger function may be adaptive, learning from each processed time series and the cost incurred by the corresponding decision. A key aspect of online adaptation is the \emph{update time}, defined as the moment at which feedback becomes available to update the trigger. Depending on the algorithm and the information accessible after a decision, three update regimes can be distinguished: (\textit{i}) \emph{Delayed updates} correspond to settings where the learning signal is only available at time $T$, after the full time series has been observed. This occurs when the outcome associated with a decision can only be assessed once the entire process has unfolded.  
(\textit{ii}) \emph{Instant updates} refer to settings where the cost of the decision is immediately observable, allowing the trigger to be updated right after stopping. This regime naturally arises in RL approaches.  
Finally, (\textit{iii}) \emph{No updates} denotes methods in which the trigger is not adapted from feedback, but instead relies on explicit cost information specified at deployment time.

\subsection{Delayed Updates}

{In standard ECTS, training typically assumes access to complete time series up to the deadline $T$. Several methods explicitly exploit full-horizon information to construct their learning signal. When deployed online, such approaches still require observing the entire time series and realized costs to form new examples to learn from. This update regime is therefore applicable only when the continuation of the series remains observable and is not affected by the trigger decision.}

{\textit{Proba-Threshold}~\cite{renaultearly} is a separable baseline that triggers a prediction when the maximum posterior probability, $p(y|\mathbf{x}_{t})$, exceeds a learned threshold.} 
An online variant maintains a set of candidate thresholds, each characterized by an empirical estimate of the average loss observed when using it. During deployment, decisions are made by selecting the decision threshold that minimizes this estimated average loss. Another variant exploits a decay parameter in order to forget previous observations more quickly: the \textit{decay-Proba-Threshold} variant includes such a mechanism (see Section \ref{sec_experimental_protocol}). Higher thresholds generally lead to later stopping times, and evaluating their associated loss may require access to the full time series.





{\textit{Calimera}~\cite{bilski2023calimera} is an anticipation-based ECTS method that estimates the difference between cost at \textit{current} time $t$ and the \textit{best reachable future cost} from the next timestep $t+1$ until the end of the time series $T$. As soon as this estimate is positive, the decision is triggered, indicating that the minimum cost is being exceeded. It relies on an ensemble of regressors learned by backward induction. We introduce a variant, referred to as \emph{Deep-Calimera}, which replaces the original ensemble of regressors with a single deep regression model, making it more suitable for online updates.}
This architectural choice has direct implications for target construction. Unlike the original \textit{Calimera} method, where the regression targets are defined iteratively from future model \textit{predictions}, \textit{Deep-Calimera} {rather exploits \textit{realized} costs in order to compute its targets (see line 8 of Algorithm \ref{algo_calimera})}.
This model takes as input five features derived from the classifier output plus the time index $t$ (i.e., the same as \textit{Alert} \cite{renault2025deep}).
\textit{Deep-Calimera} thus requires access to full time series and associated costs.

\begin{algorithm}[!htb]
\small
\KwInput{$(\mathbf{x}_{T}, y)$, $\{p(y|\mathbf{x}_{t})\}_{(1 \leq t \leq T)}$, $\mathrm{C}_m^{(u)}$, $\mathrm{C}_d^{(u)}$, DNN $f$}
\setstretch{1}

$D \gets \{\}$ \\
\tcc{generate labeled examples}
$\hat{y}_{T} \gets \argmax_{k \in \mathcal{Y}} p(k|\mathbf{x}_{T})$ \\
$A_{T} \gets p(y|\mathbf{x}_{T}) \cdot \mathrm{C}_m^{(u)}(\hat{y}_T|.) + \mathrm{C}_d^{(u)}(T)$, $\:y'_T \gets 0$ \\
\For{$t=T-1$ \KwTo $1$}{
  $\hat{y}_{t} \gets \argmax_{k \in \mathcal{Y}} p(k|\mathbf{x}_{t})$ \\
  $A_t \gets p(y|\mathbf{x}_{t}) \cdot \mathrm{C}_m^{(u)}(\hat{y}_t|.) + \mathrm{C}_d^{(u)}(t)$ \\
  $y_t' \gets \mathbbm{1}(y'_{t+1} < 0) y'_{t+1} + A_{t+1} - A_t$ \\
  $X'_t \gets$ \texttt{generate\_features}($p(y|\mathbf{x}_{t}$), $t$) \\ 
  $D \gets D \cup \{X'_t, y'_t\}$ \\
}
\tcc{update $f$ on $D$}
\texttt{minimize}($\sum_{t=1}^{T-1} ||f(X'_t) - y'_t ||^2$)
\caption{\textit{Deep-Calimera} update algorithm}
\label{algo_calimera}
\end{algorithm}

\vspace{-4mm}

\subsection{Instant Updates}

Instant-update approaches adjust the triggering model immediately after a decision, without observing the remaining part of the time series. They are therefore applicable when subsequent measurements are unobserved or influenced by the trigger decision itself. RL-based algorithms are naturally suited to this setting \cite{sutton2018reinforcement}.

{MAB models have been used as triggering mechanisms in early-exit neural networks~\cite{bajpaibeyond}. We adopt a similar construction by defining each arm as a candidate confidence threshold, yielding a bandit analogue of \textit{Proba-Threshold}. The \textit{UCB1} algorithm~\cite{auer2002finite} balances exploration and exploitation across thresholds values. For each processed time series, the selected threshold determines the stopping time and the incurred deployment-time loss, which serves as bandit feedback. {Another variant, \textit{SW-UCB1} \cite{garivier2008upper}, forces the trigger to forget past observations by processing only the ones contained in a sliding-window of fixed size.} Furthermore, the triggers training examples are used here to initialize the average arms rewards values, before updating online over deployment examples. Indeed, it has been shown that UCB-type algorithms could effectively benefit from offline history to initialize bandit statistics, as done in the \textit{HUCB1} method \cite{shivaswamy2012multi}. Algorithm \ref{algo_bandit} describes the \textit{HUCB1} update function.}

\begin{algorithm}[!htb]
\small
\KwInput{$(\mathbf{x}_{T}, y)$, $\{p(y|\mathbf{x}_{t})\}_{(1 \leq t \leq T)}$, $\mathrm{C}_m^{(u)}$, $\mathrm{C}_d^{(u)}$, $c \in \mathbb{R}^{+}$,  arms (decision thresholds) $K = \{ \delta_1, \cdots, \delta_I\}$, per-arms average rewards $\Bar{R} = \{ \Bar{r}_1, \cdots , \Bar{r}_I \}$, per-arms examples counts $n = \{ n_1, \cdots, n_I \}$, number of training examples $N$}
\setstretch{1}

$u \gets \sum_{i=1}^{I} n_i$, $\:\hat{t} \gets T$, $\:t \gets 1$\\
\tcc{select arm and get trigger time}
$\hat{k} \gets  \argmax_{1 \leq i \leq I} \left(\Bar{r}_i + c\sqrt{\frac{2\ln(N + u)}{N+n_i}} \right)$ \\

\While{$t \leq \hat{t}$}{
    $a_t \gets \mathbbm{1}(\argmax_{k\in \mathcal{Y}} p(k|\mathbf{x}_{t}) > \delta_{\hat{k}})$ \\
    \uIf{$a_t = 1$}{ 
        $\hat{t} \gets t$
    }
    \uElse{
        $t \gets t + 1$
    }
}
$\hat{y}_{\hat{t}} \gets \argmax_{k \in \mathcal{Y}} p(k|\mathbf{x}_{\hat{t}})$ \\
$r_{\hat{k}} \gets$ \texttt{get\_reward}($\hat{y}_{\hat{t}}, y, \hat{t}, a_{\hat{t}}$) \\

$n_{\hat{k}} \gets n_{\hat{k}} +1$ \tcp*{update arm statistics}
$\Bar{r}_{\hat{k}} \gets \Bar{r}_{\hat{k}} + \frac{\Bar{r}_{\hat{k}}- r_{\hat{k}}}{N + u} $ \\

\caption{\textit{HUCB1} update algorithm}
\label{algo_bandit}
\end{algorithm}

{\textit{Alert} \cite{renault2025deep} is a DQN-based trigger model with a separable design, trained offline. 
In this work, an online variant is obtained by enabling the exploration parameter $\epsilon$ during deployment and continuing to update the Q-network based on temporal-difference errors (see Algorithm \ref{algo_alert}). Each episode corresponds to time series processed until a decision at time $\hat{t}$ or when the deadline $T$ is reached, and the cumulative time-dependent loss defines the reward signal. This enables the \textit{Alert} to track evolving deployment-time costs using only partial time series.}

\SetKwRepeat{Do}{do}{while}

\begin{algorithm}[!htb]
\small
\KwInput{$(\mathbf{x}_{T}, y)$, $\{p(y|\mathbf{x}_{t})\}_{(1 \leq t \leq T)}$, $\mathrm{C}_m^{(u)}$, $\mathrm{C}_d^{(u)}$, $\epsilon \in [0,1]$, $\gamma \in [0, 1]$, DQN $Q$}
\setstretch{1}

$D \gets \{\}$, $\hat{t} \gets T$, $t \gets 1$ \\
\tcc{generate transitions}
\While{$t \leq \hat{t}$}{
  $X'_t \gets$ \texttt{generate\_features}($p(y|\mathbf{x}_{t}$), $t$) \\ 
  $a_t \gets$ $\left\{\begin{array}{ll}
       \argmax_{a} Q(X'_t, a) \text{ with proba. } 1-\epsilon  \\
       \sim \text{Bernoulli}(0.5) \text{ with proba. } \epsilon
  \end{array}\right.$ \\
  $\hat{y}_{t} \gets \argmax_{k \in \mathcal{Y}} p(k|\mathbf{x}_{t})$ \\
  $r_t \gets$ \texttt{get\_reward}($\hat{y}_t, y, t, a_t$) \\
  \uIf{$a_t = 1$}{
      $\hat{t} \gets t$, $X'_{t+1} \gets \emptyset$
  }
  \uElse{
    $X'_{t+1} \gets $ \texttt{generate\_features}($p(y|\mathbf{x}_{t+1}$),$t+1$) \\
    $t \gets t+1$
  }
  $D \gets D \cup \{X'_t, a_t, r_t, X'_{t+1} \}$ \\
}
\tcc{update $Q$ on $D$}
\texttt{minimize}($\sum_{t=1}^{\hat{t}} || Y_{t} - Q(X'_t, a_t) ||^2$) \\
with $Y_{t} = r_t + \gamma \max_a Q(X'_{t+1}, a)$
\caption{\textit{Alert} update algorithm}
\label{algo_alert}
\end{algorithm}

\subsection{No Updates (Cost-Agnostic Training)}

{A final category consists of methods that do not adapt their triggering model based on deployment-time feedback, but instead rely on explicit specification of costs at deployment.}

{\textit{Economy}-$\gamma$~\cite{achenchabe2021early} is a non-myopic ECTS method that estimates expected future costs and triggers a decision when the current expected cost is minimal over the remaining time horizon. The triggering model is learned in a cost-agnostic manner and exploits the misclassification cost matrix and the delay cost function as an input at deployment-time. This separation allows the method to naturally handle cost drift, provided that updated cost information is available at inference time (with a one-step lag in this work).}

{A key assumption of \textit{Economy-$\gamma$} is access to detailed cost information {at deployment-time}, 
rather than a single aggregated signal such as a realized time-dependent loss. While this enables cost anticipation, it also restricts the range of scenarios in which the method can be directly applied.}

\medskip
\noindent
{\textbf{Reproducibility:} all the approaches presented in this paper are available as open-source code at \href{https://github.com/aurelien-renault/ects_cost_drift}{Github}.}

\section{Experimental Protocol}
\label{sec_experimental_protocol}

This section describes the experimental protocol used to evaluate the proposed online ECTS methods under cost non-stationarity. We detail the  evaluation metrics, the cost drift scenarios, the data generation process, and the evaluated methods. All these elements are fixed prior to the analysis of results presented in Section~\ref{sec_results}.

\subsection{Evaluation Metrics}

{All online methods are evaluated using an \emph{eval-then-update} (prequential) protocol~\cite{10.1145/1557019.1557060}. At each deployment step $u$, a time series and its associated deployment-time costs are first used to evaluate the current model. The observed loss is then used to update the trigger according to its update strategy.} 

{
To study the impact of different trade-offs between misclassification and delay costs, we use the standard weighted formulation of the average cost:
\begin{equation}
    \textit{AvgCost}_{\alpha} = \frac{1}{N} \sum_{i=1}^{N} \alpha\, \mathrm{C}_m(\hat{y}_i \mid y_i) + (1-\alpha)\, \mathrm{C}_d(\hat{t}_i),
    \label{eq_avgcost_weight}
\end{equation}
where $\alpha \in [0,1]$ controls the relative importance of accuracy and earliness.}
{
Throughout this work, the same cost functions as \cite{renaultearly, bilski2023calimera, achenchabe2021early} are used: 
$C_m(\hat{y}| y) = \mathbbm{1}(\hat{y} \neq y), \:\: C_d(t) = \frac{t}{T}$.
}

In the presence of cost non-stationarity, two complementary evaluation metrics are considered. First, we measure the cumulative regret over $U$ deployment steps,
\[
    R_U = \sum_{u=1}^{U} \mathcal{L}^{(u)}(y, \hat{y}, \hat{t}) - \mathcal{L}^{(u)}(y, \hat{y}, t^{*}),
\]
which quantifies how each method adapts to changing costs. Second, we report $\textit{AvgCost}_{\alpha}$ computed on a separate hold-out set with drifted costs, while keeping the trigger model learned at $u$ frozen during this evaluation. This latter metric provides an interpretable and robust performance estimate under a fixed data distribution and is applicable to supervised and RL-based approaches. Bandit-based methods, though, are primarily compared using regret-based metrics, as they inherently rely on continued exploration. 


\subsection{Cost Drift Scenarios}

{To evaluate the robustness of online ECTS methods to non-stationary costs, we consider controlled cost variation scenarios affecting both misclassification and delay costs. Cost non-stationarity occurs across deployment steps $u$, while the time index $t \in [0,T]$ remains associated with the progression within an individual time series.}

{We first distinguish two temporal patterns of cost variation:
\begin{itemize}
    \item \textbf{Abrupt changes (AC)}, where costs switch suddenly from the training regime to the deployment one.
    \item \textbf{Periodic variations (PV)}, where costs evolve smoothly over deployment and return to the initial training regime.
\end{itemize}
}

{Independently of their temporal pattern, we consider two types of cost non-stationarity:
\begin{itemize}
    \item \textbf{Drift (D)}, where deployment costs differ from training ones and evolve across deployment steps $u$. 
    \item \textbf{Stochasticity (S)}, where the cost incurred for each processed time series is a realization of an underlying generative process, potentially varying across deployment steps $u$. In this case, the costs used during training represent partial information about this process (e.g., its mode or expectation), and the variability of realized costs is only observed at deployment.
\end{itemize}
}

{Combining these two dimensions yields four experimental scenarios:
\begin{itemize}
    \item \textbf{AC\_D}: abrupt deterministic cost drift: the $\alpha$ parameter drops from 0.8 in training to 0.4 during deployment.
    \item \textbf{PV\_D}: periodic deterministic cost drift: the $\alpha$ parameter goes from 1 to 0.1 and goes back to 1.
    \item \textbf{AC\_S}: abrupt stochastic cost variations: costs associated to a subgroup of classes are sampled from a log-normal distribution with mode equal to 1 and scale equal to 5, clipped between 0 and 500.
    \item \textbf{PV\_S}: same as above with mode equal to 1 and scale parameter evolving from 0.25 to 10 to 0.25 again. 
\end{itemize}
}


\subsection{Data Generation}


{Experiments are conducted on the MNIST-1D synthetic dataset \cite{greydanus2020scaling}, a ten-class classification problem based on univariate time series of fixed length $T=40$. Each generated time series contains a class-specific pattern representing the shape of a digit, embedded at a random position.
As the pattern is randomly located within the time series, classification performance improves, on average, as more time steps are observed, which is a necessary condition for ECTS~\cite{renaultearly}.} {A total of 20{,}000 time series are generated using the default parameters.
Among them, 25\% are used for the training phase (classifier, calibrator, trigger model), 50\% are reserved for deployment-time evaluation, and the remaining 25\% form the hold-out set.}

\subsection{Evaluated Methods}

{
Our experimental study compares the online ECTS triggering strategies introduced in Section \ref{sec_ects_online}, namely \textit{Proba-Threshold}, \textit{decay-Proba-Threshold}, \textit{HUCB1}, \textit{SW-HUCB1}, \textit{Deep-Calimera}, \textit{Alert}, and \textit{Economy-$\gamma$}. In addition, two static baselines are included to provide reference points for evaluating the benefit of online adaptation: (\textit{i}) \textbf{$no\_adapt$}: a static \textit{Proba-Threshold} model that does not adapt at deployment, consistently applying the threshold learned during training. (\textit{ii}) \textbf{$silver$}: a \textit{Proba-Threshold} model retrained from scratch at each deployment step $u$ using the true deployment-time cost configuration. 
}

\subsection{Implementation details}

{The classification module consists of an ensemble of 20 MiniROCKET estimators \cite{dempster2021minirocket}, i.e. one classifier per $5\%$ of the series, followed by a calibration step, based on isotonic regression.} {During deployment, incoming time series are processed in batches of size 16. The \textit{Deep-Calimera} trigger is implemented as a two-layer fully connected neural network with Leaky ReLU activations and a hidden dimension of size 64. For both \textit{Alert} and \textit{Deep-Calimera}, the deployment-time learning rate is set to $10^{-3}$.} {Unless stated otherwise, all hyperparameters are kept identical to those reported in the original papers. 
For \textit{Proba-Threshold}, an exponential decay with parameter $\gamma = 0.01$ is applied to weight recent observations more strongly. For \textit{HUCB1}, we include the \textit{SW-HUCB1} variant~\cite{garivier2008upper}, which computes bandit statistics over a sliding window of fixed size (set to 1000 in our experiments).}

\section{Results}
\label{sec_results}

In this section, we analyze the ECTS online performances under the two types of cost non-stationarity presented in Section \ref{sec_experimental_protocol}.

\subsection{Cost balance drift (\textbf{D})}

\setcounter{figure}{0}
\begin{figure}[!htb]
    \centering
    \subfloat[\textbf{AC\_D}: Abrupt drift]{\includegraphics[width=0.5\linewidth]{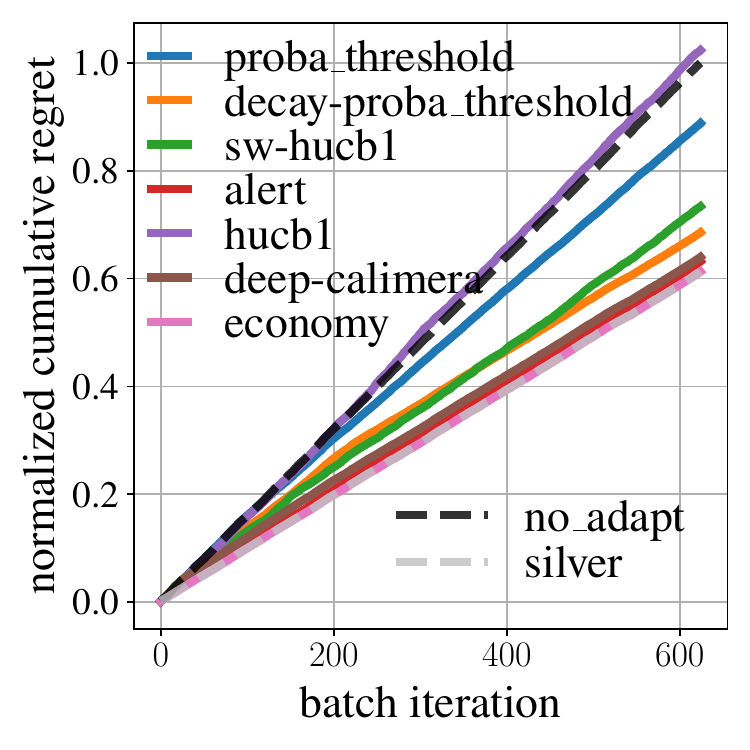}}
    \subfloat[\textbf{PV\_D}; Periodic variations]{\includegraphics[width=0.5\linewidth]{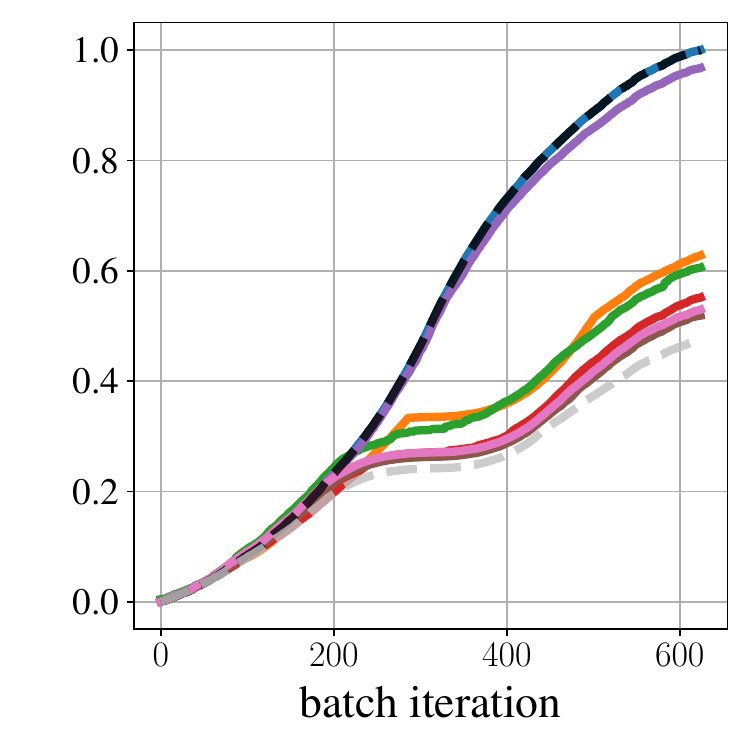}}
    \caption{Normalized cumulated regret.} 
    \label{fig:regrets}
\end{figure}


\setcounter{figure}{1}
\begin{figure*}[!t]
        \centering
    \includegraphics[width=0.8\linewidth]{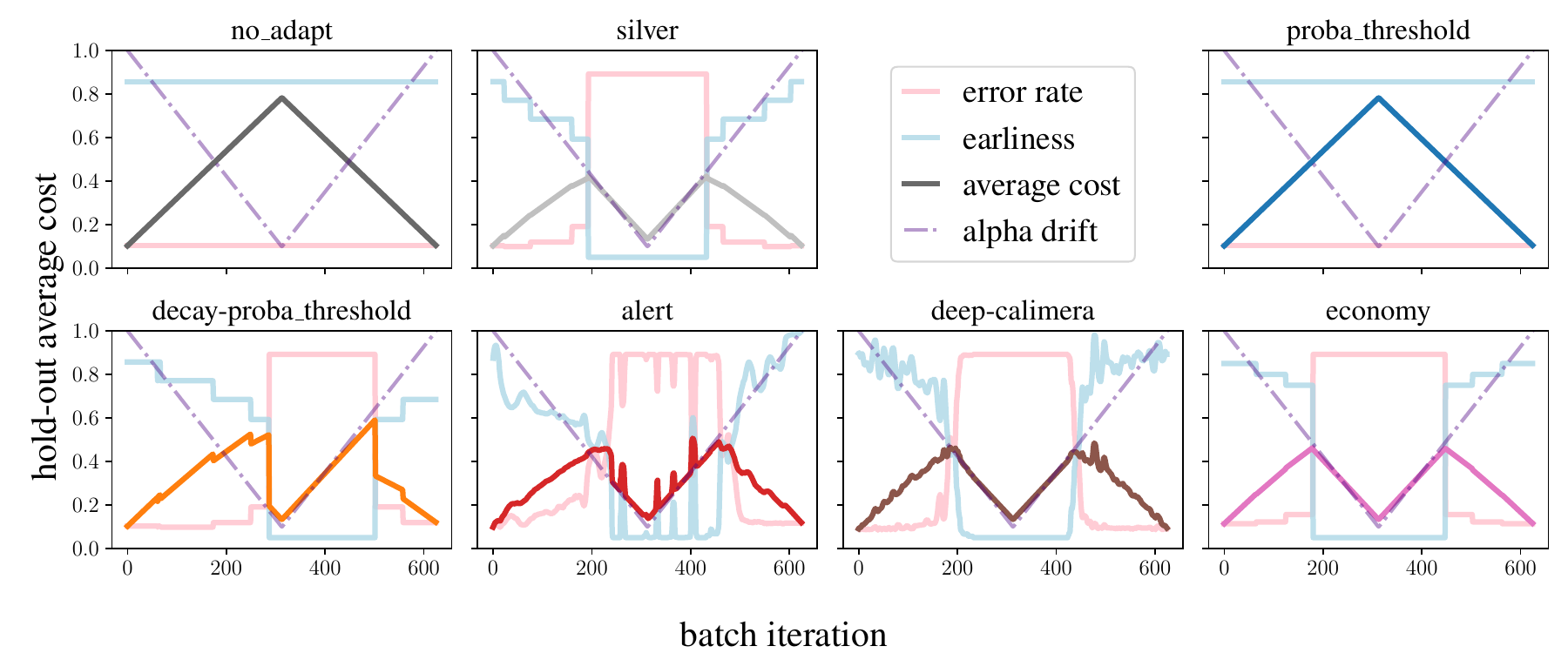}
        \caption{\textbf{PV\_D}: Hold-out \textit{AvgCost}, decomposed over both earliness and error rate axis. The thick \textit{AvgCost} line corresponds to the $\alpha$ weighted sum of earliness and error rate. Purple curve describe the evolution of the $\alpha$ parameter.}
        \label{fig:cost_gradual}
\end{figure*} 

In this subsection, we evaluate two types of $\alpha$ drift: (\textit{i}) abrupt drift (\textbf{AC\_D}) and (\textit{ii}) periodic variations drift (\textbf{PV\_D}). The regret analysis reported in Figure~\ref{fig:regrets} shows that most ECTS methods are able to adapt to both drift scenarios, with the notable exceptions of base \textit{Proba-Threshold} and \textit{HUCB1}. These results highlight the importance of forgetting mechanisms for handling non-stationarity: methods incorporating such mechanisms, namely \textit{decay-Proba-Threshold} and \textit{SW-HUCB1}, substantially outperform their full-memory counterparts. Among the approaches closest to the \textit{silver} baseline in terms of cumulative regret, which is not applicable in practice, \textit{Economy}, \textit{Alert}, and \textit{Deep-Calimera} achieve the best performance. In particular, \textit{Economy} performs especially well in this setting, where cost matrices exhibit limited variation between deployment steps. Finally, the strong performance of \textit{Alert} is particularly encouraging, as being an \textit{instant-updates} method, it relies on less information than \textit{Deep-Calimera}, which delays its updates.




Figure~\ref{fig:cost_gradual} shows the decomposition of the hold-out \textit{AvgCost} into its temporal (earliness) and misclassification (error rate) components for the \textbf{PV\_D} experiment. As $\alpha$ decreases, corresponding to increasing temporal pressure, most methods adapt their triggering behavior, as evidenced by the downward trend of the earliness (light-blue) curve, indicating that decisions are made earlier, on average. This shift is accompanied by an increase in the error rate, reflecting the higher uncertainty associated with earlier predictions. When $\alpha$ is restored to its initial value, most methods revert to their original operating regime; however, \textit{decay-Proba-Threshold} exhibits a slower recovery and requires additional time to return to its initial behavior, as its memory of past events gives it inertia.


\subsection{Stochastic costs (\textbf{S})}


The second form of cost non-stationarity considered in this work concerns stochastic cost realizations. This experiment evaluates the robustness of ECTS methods to noise and their ability to adapt under stochastic perturbations. The regret curves reported in Figure~\ref{fig:stochastic_regrets} show that, for both \textbf{AC\_S} and \textbf{PV\_S} scenarios, threshold-based methods achieve comparable performance levels, similar to that of \textit{Economy}. However, these approaches are consistently outperformed by non-myopic models, namely \textit{Alert} and \textit{Calimera}. While \textit{Deep-Calimera} attains lower cumulative regret than \textit{Alert}, this improvement comes at the cost of a more restrictive setting, as it requires access to the complete time series for updates. In contrast, \textit{Alert} can update its model online and still demonstrates strong performance. Although \textit{Economy} is also non-myopic, its performance degrades in this scenario due to the lack of temporal continuity in the cost values, which are independently drawn at each iteration. Consequently, two consecutive cost functions may differ substantially, a setting for which \textit{Economy} is not designed. Finally, bandit-based methods are adversely affected by the iterative reward normalization based on the current sampled maximum cost, which is required to satisfy boundedness assumptions but results in non-homogeneous updates.

\setcounter{figure}{2}
\begin{figure}[!htb]
    \centering
    \subfloat[\textbf{AC\_S}: Abrupt Stochastic]{\includegraphics[width=0.5\linewidth]{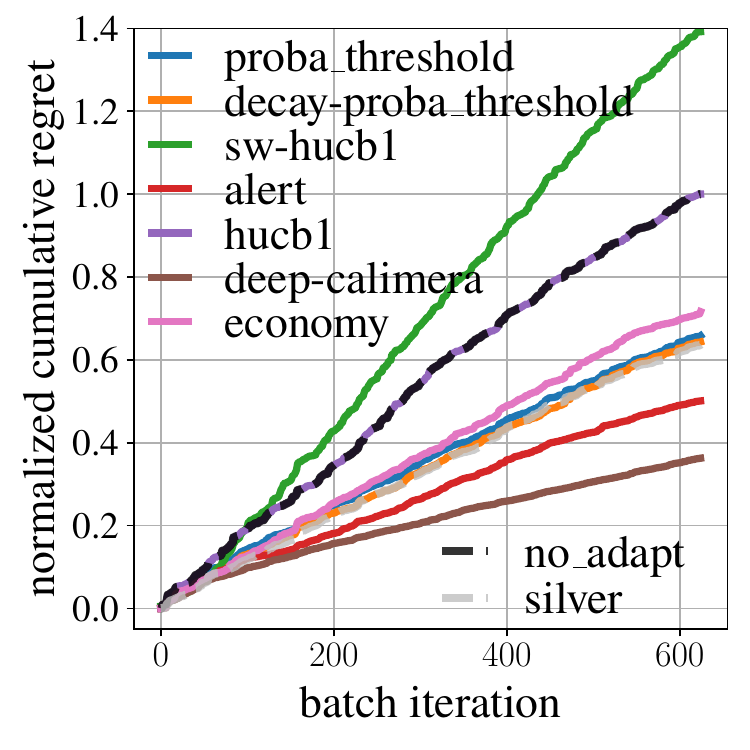}}
    \subfloat[\textbf{PV\_S}: Periodic variations Stochastic]{\includegraphics[width=0.5\linewidth]{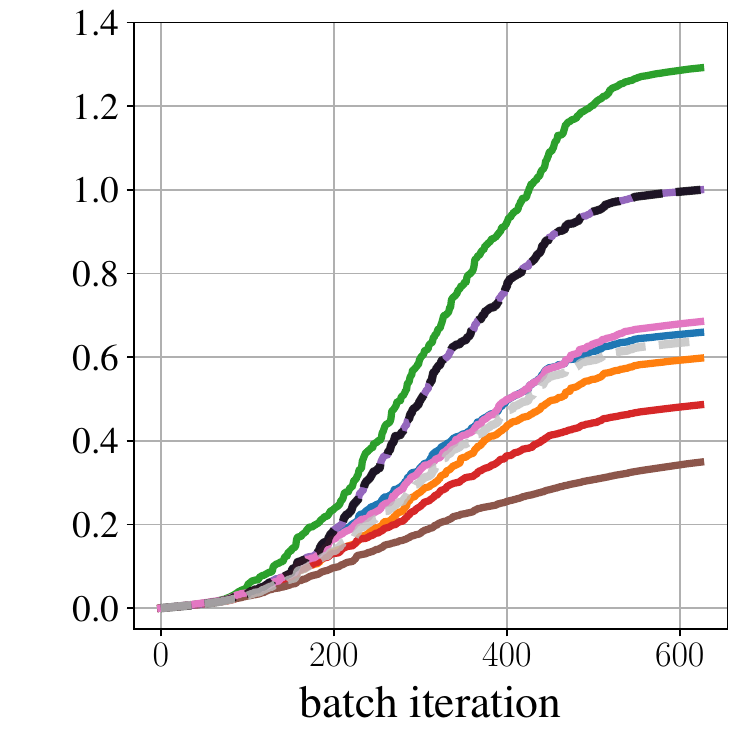}}
    \caption{Normalized cumulated regret. Costs of classes $\{1, 4, 7 \}$ (straight digits) are stochastic. Similar results were obtained with others noised classes. Averaged over five random seeds.}
    \label{fig:stochastic_regrets}
\end{figure}

\subsection{Computation time}

As ECTS methods are intended for real-time deployment, the runtimes of both inference and trigger update procedures are reported in Table~\ref{tab:times}, measured under the same experimental conditions as in \textbf{PV\_S}. As expected, \textit{Alert} and \textit{Deep-Calimera} incur higher computational overhead, whereas bandit-based and threshold-based methods exhibit lower runtimes.

\begin{table}[!htb]
    \small
    \centering
    \begin{tabular}{c|cc|c}
         Trigger & Inference & Update & Total $\downarrow$  \\
         \hline
         \textit{HUCB1} & 0.079 & 0.374 & 0.453 \\
         \textit{Economy} & 0.490 & 0 & 0.490 \\
         \textit{SW-HUCB1} & 0.119 & 0.970 & 1.089 \\
         \textit{(decay-)Proba-Threshold} & 0.076 & 2.546 & 2.622 \\ 
         \textit{Alert} & 0.780 & 5.657 & 6.437 \\
         \textit{Deep-Calimera} & 0.707 & 8.026 & 8.733 \\
    \end{tabular}
    \caption{Mean inference and update time per batch (\textit{ms}).}
    \label{tab:times}
\end{table}

\section{Conclusion}


This paper addresses ECTS under deployment-time cost non-stationarity, a practically relevant setting that is largely overlooked in existing work.   
%
Beyond the specific ECTS setting, this work contributes to the broader problem of early and adaptive decision-making under evolving cost constraints.

{We investigated two complementary sources of non-stationarity: deterministic drift in misclassification and delay costs, and stochastic deployment-time cost realizations. Focusing on separable ECTS approaches, we proposed online adaptations of representative triggering strategies that update only the trigger during deployment. 

{Our contributions include online extensions of several anticipation, bandit, and RL–based triggers, together with controlled experimental protocols explicitly modeling cost drift and stochasticity. 
In particular, RL–based triggers show strong and stable performance across all scenarios, while non-myopic methods such as \textit{Deep-Calimera} achieve lower costs when delayed updates are feasible, thus at the expense of more restrictive applicability.}

{This work opens several directions for future research, including evaluation on real-world datasets, improving end-to-end ECTS methods to make it competitive, and jointly addressing cost non-stationarity, data distribution shifts and concept drifts. 
To facilitate further research, all experiments are fully reproducible and are released as open source. 
We believe that explicitly accounting for cost non-stationarity is a necessary step toward deploying reliable and adaptive neural decision systems in real-world, long-running applications.

\bibliographystyle{IEEEtran}
\bibliography{biblio}

 \end{document}